\documentclass[conference]{IEEEtran}
\IEEEoverridecommandlockouts

\usepackage{cite}
\usepackage{amsmath,amssymb,amsfonts}
\usepackage{algorithmic}
\usepackage{graphicx}
\usepackage{textcomp}
\usepackage{xcolor}

\usepackage{hyperref} % For clickable links
\usepackage{float} 

\makeatletter
\newcommand{\linebreakand}{%
  \end{@IEEEauthorhalign}
  \hfill\mbox{}\par
  \mbox{}\hfill\begin{@IEEEauthorhalign}
}
\makeatother

\def\BibTeX{{\rm B\kern-.05em{\sc i\kern-.025em b}\kern-.08em
    T\kern-.1667em\lower.7ex\hbox{E}\kern-.125emX}}
\begin{document}

\title{
Enhancing Time Series Classification with Diversity-Driven Neural Network Ensembles
}
\author{
    \IEEEauthorblockN{Javidan Abdullayev}
    \IEEEauthorblockA{
        \textit{Université de Haute Alsace} \\
        \textit{IRIMAS}\\
        Mulhouse, France \\
        javidan.abdullayev@uha.fr}
    \and
    
    \IEEEauthorblockN{Maxime Devanne}
    \IEEEauthorblockA{
        \textit{Université de Haute Alsace} \\
        \textit{IRIMAS}\\
        Mulhouse, France \\
        maxime.devanne@uha.fr}
    \and

    \IEEEauthorblockN{Cyril	Meyer}
    \IEEEauthorblockA{
        \textit{Université de Haute Alsace} \\
        \textit{IRIMAS}\\
        Mulhouse, France \\
        cyril.meyer@uha.fr}
    \and

    \IEEEauthorblockN{Ali Ismail-Fawaz}
    \IEEEauthorblockA{
        \textit{Université de Haute Alsace} \\
        \textit{IRIMAS}\\
        Mulhouse, France \\
        ali-el-hadi.ismail-fawaz@uha.fr}
    \linebreakand

    \IEEEauthorblockN{Jonathan Weber}
    \IEEEauthorblockA{
        \textit{Université de Haute Alsace} \\
        \textit{IRIMAS}\\
        Mulhouse, France \\
        jonathan.weber@uha.fr}        
    \and

    \IEEEauthorblockN{Germain Forestier}
    \IEEEauthorblockA{
        \textit{Université de Haute Alsace} \\
        \textit{IRIMAS}\\
        Mulhouse, France \\
        \textit{Monash University, DSAI} \\
        Melbourne, Australia \\        
        germain.forestier@uha.fr}
    \and
}

\maketitle

\begin{abstract}
Ensemble methods have played a crucial role in achieving state-of-the-art (SOTA) performance across various machine learning tasks by leveraging the diversity of features learned by individual models. 
In Time Series Classification (TSC), ensembles have proven highly effective whether based on neural networks (NNs) or traditional methods like HIVE-COTE. 
However most existing NN-based ensemble methods for TSC train multiple models with identical architectures and configurations.
These ensembles aggregate predictions without explicitly promoting diversity which often leads to redundant feature representations and limits the benefits of ensembling. 
In this work, we introduce a diversity-driven ensemble learning framework that explicitly encourages feature diversity among neural network ensemble members.
Our approach employs a decorrelated learning strategy using a feature orthogonality loss applied directly to the learned feature representations.
This ensures that each model in the ensemble captures complementary rather than redundant information.
We evaluate our framework on 128 datasets from the UCR archive and show that it achieves SOTA performance with fewer models.
This makes our method both efficient and scalable compared to conventional NN-based ensemble approaches.
\end{abstract}


\section{Introduction}
Time Series Classification (TSC) is a fundamental problem in machine learning with applications across various domains including healthcare~\cite{rajkomar2018scalable}, human activity recognition~\cite{nweke2018deep}, social security~\cite{yi2018integrated}, remote sensing~\cite{pelletier2019temporal}, etc..
Increasing availability of large-scale time series datasets, such as the UCR archive~\cite{dau2019ucr}, has led to the development of more advanced classification methods. 
Recent progress in deep learning has significantly improved TSC performance by leveraging convolutional neural networks (CNNs) which can automatically extract meaningful temporal features~\cite{middlehurst2024bake}.

Currently state-of-the-art (SOTA) deep learning models for TSC including InceptionTime\cite{ismail2020inceptiontime}, H-InceptionTime\cite{ismail2022deep} and LITETime~\cite{ismail2023lite} achieve high accuracy using ensemble learning. 
The \textit{Time} suffix in their names signifies that these models are ensembles of five identical neural networks, each trained independently. 
These methods combine predictions from their individual members to improve classification performance.
However they do not explicitly enforce feature diversity within the ensemble just relying instead on random initialization to introduce variation. 
This often leads to feature redundancy and limits the potential gains from ensembling.

In ensemble learning, diversity among individual models is key for improving generalization\cite{dietterich2000ensemble}. 
Traditional approaches such as HIVE-COTE\cite{middlehurst2021hive} achieve diversity by ensembling heterogeneous models with different architectures. 
However, in this work, we take a different approach by focusing on homogeneous neural network ensembles. Our goal is to promote diversity among models with the same architecture by encouraging them to learn distinct feature representations.
This is an important but often overlooked aspect of deep ensembles where models trained independently often converge to similar solutions.

To address this issue we propose a decorrelated learning framework that explicitly promotes feature diversity within ensembles by penalizing redundant representations. 
Inspired by knowledge distillation~\cite{ay2022kdfcn}, we introduce a feature orthogonality loss that forces ensemble models to learn complementary rather than overlapping features. 
This loss minimizes the cosine similarity between feature vectors produced by different ensemble members.
As a result, each model captures unique aspects of the input, reducing redundancy across the ensemble.
This enhances generalization without increasing computational complexity or modifying the base model architecture.

\begin{figure*}[h]
    \centering  
    \includegraphics[width=0.9\linewidth]{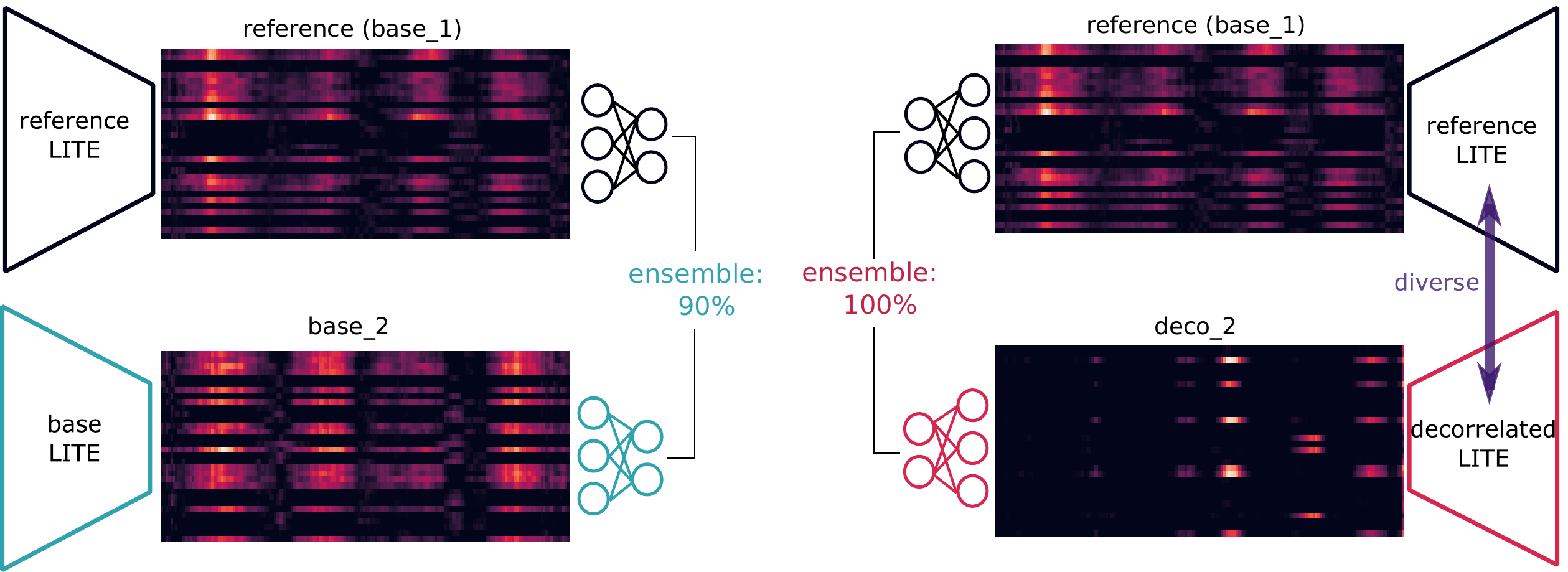}
    \caption{Comparison of ensemble model performances and feature maps on the BirdChicken dataset, from standard and decorrelated training.}
\label{fig:main_figure}
\end{figure*}

The overall idea is exemplified using the BirdChicken dataset from the UCR archive, as shown in Figure~\ref{fig:main_figure}.
On the left, two base LITE models trained separately with different initializations produce highly similar feature maps.
Conversely, in our proposal, a second decorrelated LITE model is guided to learn diverse features compared to the first base model trained previously. Figure~\ref{fig:main_figure} clearly illustrates on the right side that the decorrelated model learned different features. 
This differentiation enables the decorrelated ensemble to achieve 100\% test accuracy which exceeds the 90\% accuracy achieved by the base ensemble. 
Furthermore, it also exceeds the best performance of SOTA models such as InceptionTime~\cite{ismail2020inceptiontime} which achieves a maximum test accuracy of 95\% using an ensemble of five models.
It suggests that learning diverse features in an ensemble of deep models can result in better generalization and classification performance.

We assessed our approach on 128 datasets from the UCR archive, demonstrating its effectiveness across diverse real-world time series data. 
As a result our approach achieves performance comparable to LITETime~\cite{ismail2023lite} but with fewer models which offers a more efficient solution.

Our main contributions in this work are:
\begin{itemize}
    \item We propose a novel diversity-driven ensemble framework that explicitly promotes feature diversity and improves deep ensemble effectiveness for TSC.
    \item  We conduct a comprehensive empricial validation across 128 UCR datasets, showing that our method achieves SOTA-level performance with fewer models and improves efficiency.
    \item We provide quantitative and qualitative diversity analysis to demonstrate increased feature diversity using Fréchet Inception Distance (FID) scores and t-SNE visualizations of learned convolutional filters.
\end{itemize}

The rest of the paper is organized as follows: Section~\ref{sec:background_related_work} provides background information and discusses related work on deep learning-based TSC and ensemble methods.
Section~\ref{sec:proposed_method} describes our proposed decorrelated learning framework.
Section~\ref{sec:experimental_evaluation} presents the experimental setup, dataset details and evaluation metrics, followed by a detailed analysis of results.
Finally Section~\ref{sec:conclusion} concludes the paper and discusses potential future directions.

\section{Background and Related Work}\label{sec:background_related_work}
In this work we focus on TSC task where recent studies demonstrated that deep neural networks (DNNs) leveraging 1D temporal convolutions achieve impressive performance~\cite{mohammadi2024deep}.
Many traditional TSC algorithms especially those not based on deep learning rely heavily on feature engineering as part of the classification task.
While this approach has shown strong performance in certain cases~\cite{middlehurst2024bake} it often requires domain expertise and poses challenges in scalability and automation.
On the other hand deep learning models integrate feature extraction and classification into a single pipeline and optimize both jointly.
This capability allows them to scale efficiently to large datasets and take advantage of hardware acceleration such as GPUs.

SOTA deep learning models for TSC often employ ensemble learning where multiple models of the same architecture are trained independently with different initializations and their predictions are combined~\cite{ismail2023lite}.
While ensembling has been shown to improve accuracy~\cite{dietterich2000ensemble}, this approach does not explicitly enforce feature diversity among individual models.
As a result, ensembles may suffer from redundant feature representations which limits their overall effectiveness.

In this work, we address this issue by focusing on increasing diversity among ensemble members. 
Specifically, we introduce a decorrelated feature loss that actively encourages diverse feature learning across models during training.
Our approach optimizes ensemble diversity by ensuring that individual models learn complementary instead of redundant representations which leads to improved generalization and classification performance.

\subsection{Definitions}
A univariate time series is 
defined as an ordered set of numerical values that represents evolution of a specific quantity over time.
A time series dataset is denoted as  $\mathcal{D} = {(\mathbf{X}_i,\mathbf{Y}_i)}_{i=1}^{N}$ where $N$ represents the number of samples, $\mathbf{X}_i$ is an individual time series and $\mathbf{Y}_i$ is its corresponding label vector.
The label vector follows a one-hot encoding scheme where $ \mathbf{Y}_i  \in \mathbb{R}^C $ represents a class label $c$ from a set of $C$ predefined categories.

Time series classification (TSC) task involves assigning a class label to a given sample  based on its temporal characteristics.
The objective is to train a model that can effectively identify patterns, trends and dependencies within time series data.
Formally, the task consists of learning a mapping function $f: \mathbf{X} \rightarrow \mathbf{Y}$ that accurately classifies each input time series $\mathbf{X}_i$ into one of the predefined categories in $\mathbf{Y}_i$.

% In deep learning-based TSC, the function $f$ is typically modeled as a Deep Neural Network (DNN). 
% During trainig the model learns to map time series inputs to their correct labels by optimizing a loss function $L$ that measures the difference between the predicted labels $\mathbf{\hat{Y}}$ and the tru labels $\mathbf{Y}$.
% One commonly used loss function for classification tasks is the categorical cross-entropy loss ($L_{CE}$) which quantifies the divergence between predicted class probabilities and ground truth labels.
% The goal of training is to find the optimal function $f^*$ that minimizes $L_{CE}$ over the training dataset $\mathbf{D}_{\text{train}}$, ensuring that the model generalizes well to unseen time series samples.

\subsection{Deep Learning for Time Series Classification}
Time series analysis has been a fundamental area of research for many years with various machine learning techniques applied to tasks like TSC. 
Early approaches often relied on similarity-base methods such as Dynamic Time Warping (DTW)~\cite{dau2018optimizing} between time series or classification models such as Random Forests~\cite{lucas2019proximity} and Support Vector Machines (SVMs)~\cite{bagheri2016support}. 
A key limitation of these traditional methods is that feature extraction is typically treated as a separate process from classification which can lead to information loss and increase the complexity of the overall pipeline.
 
In recent years deep learning has gained significant popularity in the TSC due to its ability to learn complex patterns from raw time series data without requiring manual feature engineering. 
One of the first deep learning models applied to TSC was the Multi-Layer Perceptron (MLP)~\cite{ismail2019deep} but its fully connected nature made it inefficient in capturing temproal dependencies.
A major step forward came with the introduction of 1D CNNs which proved highly effective for extracting meaningful features from time series data.
The Fully Convolutional Network (FCN)~\cite{wang2017time} was one of the first deep learning models to achieve strong results in TSC. 
FCN consists of three convolutional blocks, each containing a convolutional layer, batch normalization and activation functions. 
Unlike standard CNN architectures, FCN does not use pooling layers which allows it to preserve the original time series length and retain temporal relationships more effectively.
Following FCN, researchers introduced ResNet for TSC~\cite{wang2017time} which applies nine convolutional layers but also incorporates residual connections. 
These connections improve gradient flow, mitigate information loss and make training deeper models more stable. 
In 2020, Inception-based model were introduced for TSC in the form of InceptionTime\cite{ismail2020inceptiontime}, drawing inspiration from Google's Inception v4\cite{szegedy2017inception}. 
InceptionTime consists of six Inception modules where each module applies convolutions of different kernel sizes to capture patterns at varying temporal resolutions.
Building on InceptionTime, an improved model called Hybrid Inception (H-Inception)\cite{ismail2022deep} was developed, incorporating custom convolutional filters in the initial layers to enhance feature extraction and classification performance. 

Recently, LITE (Light Inception with Boosting Techniques) was proposed as a more efficient alternative to Inception-based model\cite{ismail2023lite}. 
LITE significantly reduces parameter count to just 2.34\% of that of InceptionTime while maintaining competitive classification performance. 
The model consists of three convolutional layers, combining custom, multiplexed, dilated and depthwise separable convolutions to reduce computational cost while preserving predictive accuracy

To achieve SOTA performance, ensemble versions of these models, InceptionTime, H-InceptionTime and LITETime, are widely used in TSC. 
Each model is trained five times under the same setup but with different initializations and their predictions are combined to produce a final ensemble output. 
While ensembling improves classification accuracy these approaches do not explicitly enforce diversity among the individual classifiers. 
Instead they rely solely on random initialization to introduce variation. 
% This does not guarantee that the models will learn sufficiently different feature representations, often leading to redundant features that limit the potential benefits of ensembling.
However, this does not guarantee diverse feature learning and often leads to redundant representations.

\subsection{Ensemble Learning}
Ensemble learning is a powerful machine learning technique that improves performance by combining predictions from multiple individual models. It leverages their collective strengths to mitigate individual weaknesses and reduce generalization errors~\cite{dietterich2000ensemble}.
% Ensemble learning is a powerful machine learning technique that enhances overall performance by combining predictions from multiple individual models, leveraging their collective strengths to mitigate individual weaknesses and reduce generalization errors~\cite{dietterich2000ensemble}.
Common ensemble strategies include bagging where models are trained on different subsets of the training data~\cite{breiman1996bagging}, boosting which iteratively focuses on difficult-to-classify samples~\cite{freund1996experiments} and stacking where predictions of base models are combined through a meta-learner~\cite{wolpert1992stacked}. \\
In deep learning, ensemble approaches have been highly successful in improving generalization and reducing overfitting, particularly in complex tasks such as image recognition and time series classification. 
The work~\cite{ismail2019deep} demonstrated the effectiveness of deep neural network ensembles for time series classification by highlighting the importance of combining diverse predictions for improved accuracy.
The recent model CoCaLite~\cite{badi2024cocalite} which utilizes ensemble strategies to balance computational efficiency and predictive accuracy has demonstrated SOTA performance.
It is important to note that most existing works overlook critical aspects of feature diversity within ensembles.
Various techniques such as decorrelated learning and feature orthogonality have been introduced to explicitly promote diversity during training~\cite{lee2023importance}. 
These methods are particularly relevant for deep neural networks where similar architectures and training setups can lead to redundant feature extraction and  limits ensemble benefits. 
For instance, the kernels tailored for time series similarity have been employed with ensemble models to enhance temporal pattern recognition~\cite{lines2015time}. \\
The key advantage of ensembling lies in its ability to exploit the diversity of features learned by each individual model. 
Because of the stochastic nature of deep learning, each model may learn slightly different features which collectively enhance overall performance of the ensemble.
In convolutional neural networks this diversity is largely attributed to the variation in features learned by each model. 
Our experiments have shown that when the models in an ensemble learn very similar features, the performance gains are minimal.
%, merely reflecting the average of individual model performances (i.e., the arithmetic mean of accuracies). 
This observation is really intuitive and underscores the importance of feature diversity in the success of ensemble learning.

\subsection{Decorrelated Learning}
To the best of our knowledge, most SOTA deep learning-based TSC models rely on training multiple instances of the same model and ensembling their outputs~\cite{ismail2020inceptiontime, ismail2022deep, ismail2023lite}. 
Base models within these ensembles are trained independently without mechanisms for sharing information or coordinating their learning. 
This independent training process often leads the model to converge similar, nearby local minima which results in reduced diversity.
Both theoretical and experimental studies suggest that generalization ability of an ensemble can be greatly enhanced if base models are negatively correlated~\cite{tumer1996error}.

In contrast, decorrelated learning explicitly introduces shared training mechanisms that encourage diversity among ensemble members.
Instead of training models independently, decorrelated learning ensures that training process of each model is influenced by others in the ensemble.
By explicitly minimizing correlations between filters, features or predictions decorrelated learning ensures that each model is guided to learn distinct patterns.
This approach helps the features produced by different models to complement one another rather than overlap.
This shared training paradigm is particularly valuable in ensemble settings where diversity among individual models plays a curcial role in improving overall performance~\cite{kuncheva2003measures}. 

Techniques such as Orthogonality Loss have been developed to enforce feature decorrelation by minimizing cosine similarity between filters~\cite{yang2020orthogonality}. 
These methods are particularly relevant for tasks where redundant feature extraction limits performance, as is often observed in deep neural networks trained on datasets with identical architectures. 
Most existing approaches in CNN-based models apply orthogonality loss to convolutional filters to achieve diversity in the feature space~\cite{ayinde2019regularizing}.

In this work, we demonstrate that for TSC tasks applying orthogonality loss to convolutional filters does not lead to meaningful diversity in features. 
To address this, we propose applying orthogonality loss directly to the feature outputs. 
This approach ensures a higher level of feature diversity which is critical for improving ensemble performance. 
Additionally, decorrelation has been shown to be effective in unsupervised representation learning, helping to mitigate representational collapse and improve downstream task performance~\cite{lee2023importance}. 
These findings reinforce the value of decorrelated learning in addressing redundancy and enhancing the discriminative power of deep learning models.

Our method promotes diversity explicitly at the level of convolutional features, a strategy we have found effective for ensemble learning in TSC tasks. 
We utilize a diversity-driven auxiliary loss to improve diversity among ensemble models during training. 
The core idea of our decorrelated learning approach is to train ensemble models sequentially, ensuring that each model learns features distinct from those learned by previously trained models. 
In this paper, we focus on exploring the impact of decorrelated learning in convolution-based TSC models to deepen our understanding of its effects.

\section{Proposed Method}\label{sec:proposed_method}
In this section we introduce our diversity-driven ensemble learning framework for TSC.  
We first describe the transition from filter-level orthogonality to feature-level orthogonality, highlighting its advantages in promoting meaningful diversity among ensemble members.  
Next we provide a detailed description of our proposed framework, outlining the sequential training process and incorporation of feature orthogonality constraints.  
Finally we present the mathematical formulation of the feature orthogonality loss which serves as the core of our method.

\begin{figure*}
    \centering    
    \includegraphics[width=0.75\linewidth]{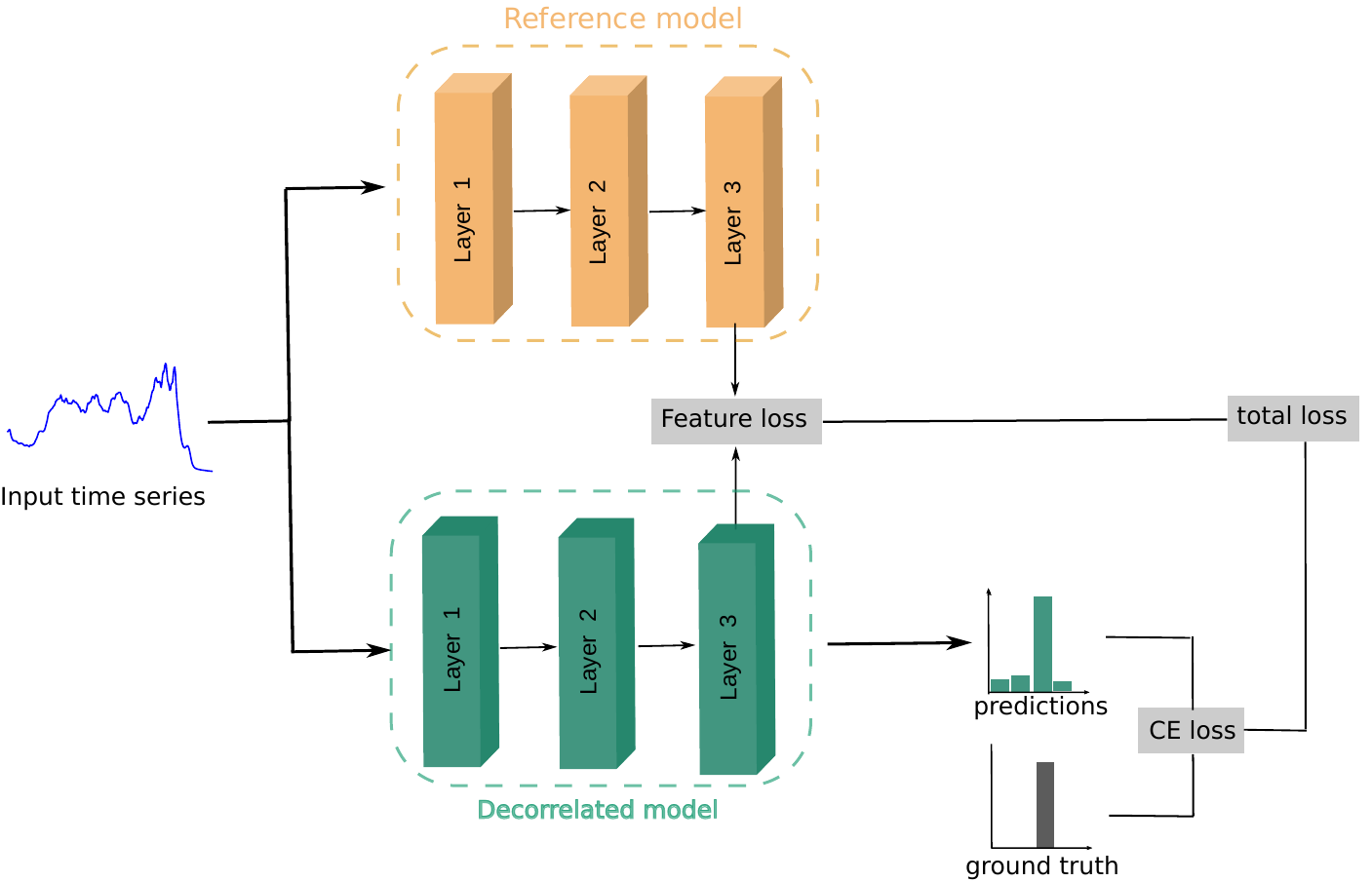}
    \caption{\centering Proposed decorrelated learning framework for a 2-model ensemble where the decorrelated model is trained with feature orthogonality loss to enhance diversity.}
    \label{fig:architecture}
\end{figure*}

\subsection{Rethinking Diversity: From Filter Orthogonality to Feature Orthogonality}
The primary objective of this work is to enhance diversity among ensemble models while maintaining the performance of individual models. 
One possible approach to promote feature diversity is to use techniques such as Deep Negative Correlation Classification (DNCC)~\cite{zhang2019nonlinear} where diversity is encouraged
during training by explicitly forcing one model to focus on samples that are misclassified or have low confidence in another model.
However, in our experiments with LITE model~\cite{ismail2023lite} we observed that it achieves 100\% training accuracy on majority of datasets from UCR archive
This indicates that the model effectively fits training data and leaves minimal room for misclassified or low-confidence  samples to enhance diversity through misclassification-based methods. 
Given this limitation we explored alternative methods for promoting diversity, focusing on feature and filter-level diversity. 
From the literature we identified several works that employ filter orthogonality loss to promote diversity among ensemble models, particularly in image classification tasks~\cite{wang2020orthogonal, yang2020orthogonality, ayinde2019regularizing}. 
The central idea behind these methods is to enforce orthogonality between convolutional filters, encouraging them to learn distinct representations. 
From our experiments we realized that filter orthogonality often shifts discriminative patterns rather than creating diverse representations. 
While filters become orthogonal, feature maps remain similar, limiting the benefits of decorrelation. 
Thus filter-level orthogonality alone does not guarantee meaningful diversity in ensembles of time series classifiers.

To address this challenge, we propose a feature diversity loss function and a diversity-driven optimization strategy aimed at encouraging each model in the ensemble to extract distinct features from the input data.
Unlike existing approaches that target filter diversity our method focuses on promoting diversity directly in the feature space itself. 
Specifically, we introduce an orthogonality loss function that operates on the feature outputs. 
The goal is to explicitly enforce orthogonality between the feature representations produced by different models in the ensemble. 
This loss function encourages each model to learn orthogonal features thereby reducing redundancy in the feature space.
By encouraging orthogonal feature representations we aim to promote diversity among individual models within the ensemble, ultimately improving the generalization and overall performance. 
% \textcolor{orange}{Orthogonal feature representations promote diversity among ensemble models which helps improve generalization and overall performance.}
This approach ensures that each model contributes unique information to the ensemble, leading to a more robust and generalizable classification system.

\subsection{Framework Overview}
In this work we introduce a sequential training framework designed to enhance diversity in neural network ensembles for TSC. 
Our approach explicitly encourages each model in the ensemble to learn distinct feature representations thereby improving generalization.
Unlike conventional ensemble methods where models are trained independently our framework introduces decorrelated models which are explicitly guided to learn features that are orthogonal to those of previously trained models.
% \jac{
% \textit{The training process follows these key steps:}
% \begin{enumerate}
%     \item \textit{Train the first model} in the ensemble using standard cross-entropy (CE) loss. This model acts as the base model.
%     \item \textit{Extract feature outputs} from the final convolutional layer of the trained model. These serve as the reference for orthogonality constraints.
%     \item \textit{Train the next model} in the ensemble (the decorrelated model) using a joint loss function that combines:
%     \begin{itemize}
%         \item \textit{Cross-entropy loss} to maintain classification accuracy, and
%         \item \textit{Feature orthogonality loss} to penalize similarity with the feature representations of previously trained models.
%     \end{itemize}
%     \item \textit{Repeat this process} for each additional model in the ensemble. At each step, the new model is trained to produce features that are orthogonal to those from \textit{all earlier models}. The orthogonality loss is averaged over these models to ensure consistency.
% \end{enumerate}
% }
The training follows a sequential process where each base model in the ensemble $(n)$ is trained to minimize feature redundancy with respect to all earlier models $(<n)$.

To achieve this goal we define an orthogonality function that measures the degree of overlap between the feature representations of the current model and those of previously trained models. 
This function penalizes feature similarity, ensuring that the newly trained model captures complementary information. 
The resulting feature orthogonality loss is combined with cross-entropy loss to form the training loss. 
By optimizing both losses jointly, our framework promotes feature diversity while maintaining high classification accuracy.

Figure~\ref{fig:architecture} illustrates the overall architecture of our proposed method, highlighting how cross-entropy (CE) loss and feature orthogonality (FO) loss are integrated into the training process. 
CE loss ensures that each model learns to accurately map time series inputs to class labels. 
Meanwhile FO loss is applied directly to the feature outputs of the final convolutional layer to encourage diversity among ensemble members.

A key consideration in DNNs is that lower layers typically learn generic features while upper layers capture task-specific representations~\cite{yosinski2014transferable}.
Applying orthogonality constraints to early layers can disrupt essential feature learning, negatively impacting performance. 
To mitigate this issue we apply FO loss only to the final layer of the LITE model, ensuring that only high-level features are decorrelated thereby maintaining both diversity and classification performance.

The total loss function balances CE loss and FO loss, ensuring that both classification accuracy and feature diversity are optimized during training. 
In ensemble configurations, this process is iteratively applied to all base models. 
For each new model FO loss is computed relative to all previously trained models and the results are averaged to ensure consistency.
During training, feature outputs from the base models are extracted and each decorrelated model is explicitly optimized to produce feature representations that are distinct from its predecessors. 
This approach enhances ensemble diversity, leading to a more robust and generalizable classification system while maintaining performance.

\subsection{Diversity Loss}
% To enhance diversity among ensemble models, we employ the cosine similarity loss which is applied directly to feature outputs. 
To enhance diversity among ensemble models, we employ a feature orthogonality loss based on cosine similarity, applied directly to the feature outputs.  
While traditional methods focus on filter-level orthogonality, we found that encouraging orthogonality in the feature space leads to more meaningful diversity in the context of time series data.
This ensures that each model in the ensemble contributes unique, complementary information.

Let $\mathbf{F}_{\text{deco}} \in \mathbb{R}^{B \times C \times T} $ and $ \mathbf{F}_{\text{base}} \in \mathbb{R}^{B \times C \times T} $ represent the feature outputs of the decorrelated and base models where $B$, $C$ and $T$ denote batch size, number of channels, and time series length, respectively. 
The diversity loss is defined as in the following equation~\ref{eq:orthogonality_loss}:

\begin{equation} 
    L_{orth} =  \sum_{i \neq j} \left| 
    \frac{\mathbf{F}_{deco, i} \cdot \mathbf{F}_{base, j}^{\top}}{|\mathbf{F}_{deco, i}| |\mathbf{F}_{base, j}|} 
    \right| 
    \label{eq:orthogonality_loss} 
\end{equation}
    
This formulation penalizes overlap between features, encouraging decorrelated model to learn distinct representations relative to the base models.
By penalizing the magnitude of off-diagonal elements in the similarity matrix, we encourage feature independence while maintaining computational efficiency.
Notably, the use of cosine similarity is not computationally expensive as matrix operations can be efficiently parallelized.
The diversity loss is then integrated into the total loss function as shown in the equation~\ref{eq:total_loss}:

\begin{equation}
    \mathcal{L}_{\text{total}} = \alpha \mathcal{L}_{\text{CE}} + (1 - \alpha) \cdot \mathcal{L}_{\text{orth}}
    \label{eq:total_loss}
\end{equation}

\( \mathcal{L}_{\text{CE}} \) is the cross-entropy loss which ensures accurate classification and \( \alpha \) is a weight parameter that balances cross-entropy and feature diversity. 
In our experiments, we set \( \alpha = 0.5 \), giving equal importance to both losses.
By minimizing this total loss, the decorrelated model learns features that are both task-relevant and orthogonal to those of previously trained models, enhancing diversity across the ensemble.

\section{Experimental Evaluation}\label{sec:experimental_evaluation}
\subsection{Experimental Setup}
\subsubsection{Data}
To validate effectiveness of our proposed approach we evaluated it on UCR Archive~\cite{dau2019ucr}, 
the largest publicly available repository for time series classification. The archive includes 128 univariate time series datasets from various domains, such as healthcare, motion tracking and sensor data, with diverse characteristics in terms of sequence length, sample size, and class distribution. The number of classes ranges from 2 to 60, providing a broad evaluation spectrum.
For a fair comparison with state-of-the-art methods we used the original train/test splits provided by the UCR Archive. All time series were z-normalized to ensure zero mean and unit variance, reducing the influence of scale differences and emphasizing intrinsic temporal patterns.

\subsubsection{Experimental protocol}
For our experiments, we followed the exact same setup as described in the LITE paper~\cite{ismail2023lite}.
We trained five standard LITE classifiers and also four decorrelted LITE classifiers which we refer to as  \textit{base} and \textit{decorrelated} models respectively. 
The \textit{base} models trained using only CE loss while the decorrelated models incorporated an additional feature diversity loss.
To ensure a robust comparison and confirm that performance changes are not due to random initialization, each base model was initializaed with a unique seed.
For consistency, decorrelated models were initlized with the same seeds as their corresponding base models.
This guarantees that any observed performance improvements are due to the feature orthogonality (FO) loss rather than randomness in initialization.

The training process for decorrelated ensembles can be summarized as follows:
\begin{enumerate}
    \item \textit{Base Model Training}: The first model in each ensemble is trained using only CE loss.
    \item \textit{Decorrelated Model Training}: Subsequent models in the ensemble are trained as decorrelated models. 
    \item \textit{Sequential Training}: For each additional model, FO loss is computed on all previously trained models from the ensemble. Specifically, the FO loss formula for (n+1)-th model is given by equation~\ref{eq:sequential_orth}.
    \begin{equation}
        L_{\text{orth}}^{n+1} = \frac{1}{n} \sum_{i=1}^{n} L_{\text{orth}}(F_{n+1}, F_{i})
        \label{eq:sequential_orth}
    \end{equation}
    where $F_{n+1}$ and $F_{i}$ represent the feature outputs of the $(n+1)$-th and $i$-th models, respectively, $L_{\text{orth}}$ denotes the orthogonality loss and $n$ is the number of previously trained models from the ensemble.
\end{enumerate}

We evaluate four ensemble cofigurations with sizes varying from two to five models and systematically comparing decorrelated ensembles against base enesembles.
Each ensemble consists of a reference model combined with either additional independently trained base models or decorrelated models trained sequentially.
For base ensembles, all models are trained independently without any form of coordinated learning and their predictions are aggregated. 
In contrast decorrelated ensembles are constructed by progressively replacing base models with their corresponding decorrelated versions which are trained sequentially. 
In this process, each decorrelated model is guided to learn feature representations that are explicitly diverse from those of previously trained models in the ensemble.
In all cases, the first base model serves as a fixed reference to ensure consistency across configurations. 
Each ensemble configuration is trained five times independently and the final results are obtained by averaging performance across these five runs.
To denote different ensemble types we use LiteTime-$N$ to represent an ensemble of $N$ independently trained base models while Deco-LiteTime-$N$ refers to an ensemble where all but the reference model are decorrelated.

\subsubsection{Comparison Protocol}
We evaluate classification performance using accuracy across the 128 datasets in the UCR archive.
To compare our decorrelated framework with standard ensembles, we use the \textit{Multi-Comparison Matrix (MCM)}~\cite{ismail2023approach}, which ranks classifiers by \textit{Mean Accuracy}—the average accuracy across all datasets. This provides a more interpretable comparison than traditional average-rank methods~\cite{benavoli2016should}.
MCM also reports the \textit{Mean Difference}, which measures the average accuracy gap between pairs of classifiers, and the \textit{Win/Tie/Loss} counts across datasets.
Statistical significance is assessed using the \textit{Wilcoxon signed-rank test}~\cite{wilcoxon1992individual} with \(p < 0.05\). Significant results are shown in bold.

\subsubsection{Implementation details}
The LITE model was used as the base architecture, following the original configuration described in prior work~\cite{ismail2023lite}. 
The models were trained using the Adam optimizer with an initial learning rate of 0.001,
a reducing factor of 0.5 and a patience of 50. 
Each model was trained for 1500 epochs with a batch size of 64.
All experiments were conducted on a system equipped with an NVIDIA RTX 4090 GPU with 24GB
of memory, running Ubuntu 22. 
The models were implemented using PyTorch 2.5.1 and Python 3.12. 
The source code is publicly available \href{https://github.com/MSD-IRIMAS/decorrelated-learning}{https://github.com/MSD-IRIMAS/decorrelated-learning}.

\subsection{Overall Performance on UCR Archive}
In this section we present a comparative analysis of our proposed framework against the SOTA LITETime~\cite{ismail2023lite}.
% and classic ensemble configurations.
Our objective is to evaluate effectiveness of our diversity-driven approach in enhancing generalization and classification accuracy by analyzing various ensemble configurations.

\begin{figure}[h!]
    \centering
    \includegraphics[width=\linewidth]{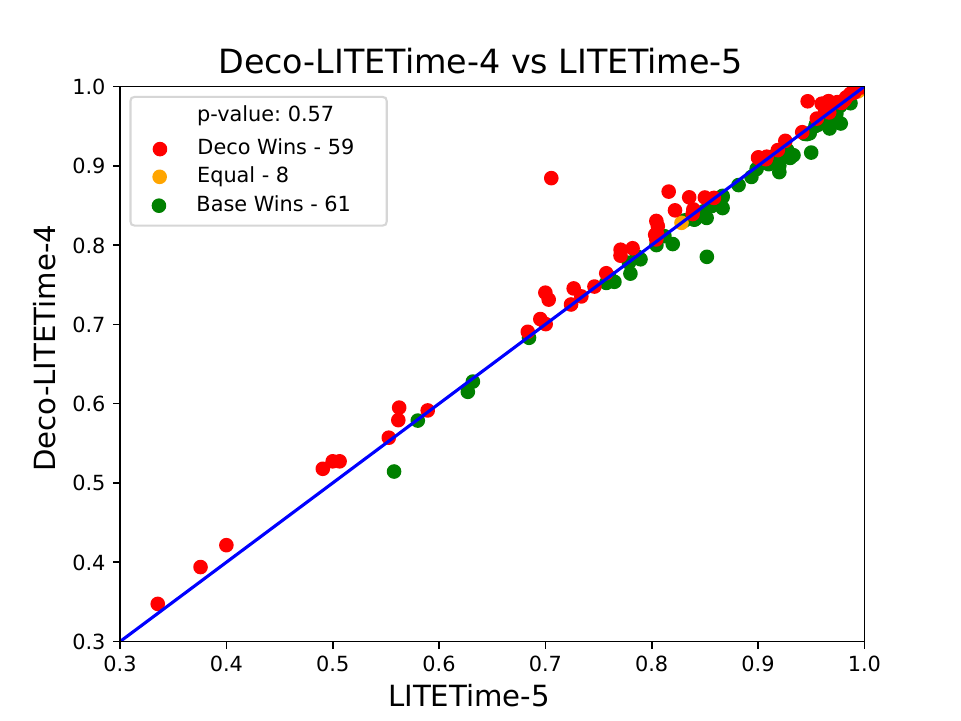}
    \caption{\centering Performance comparison between Deco-LITETime-4 and LITETime-5.}
    \label{fig:dec_ens_4_vs_litetime}
\end{figure}

\begin{figure*}
    \centering
    \includegraphics[width=0.85\textwidth]{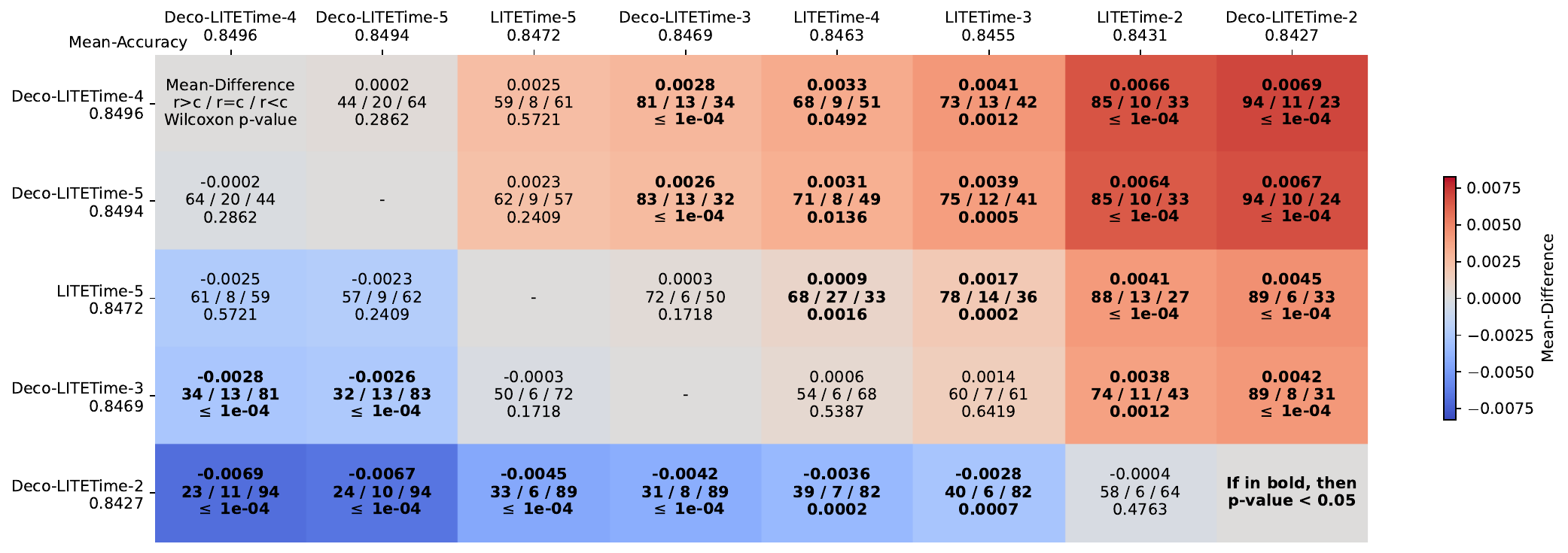}
    \caption{The Multi-Comparison Matrix illustrates performance of each decorrelated and base ensemble variants in one-vs-one comparisons.}
    \label{fig:mcm_sota} 
\end{figure*}

The comparative results are presented in the form of an MCM matrix as illustrated in Figure~\ref{fig:mcm_sota}.
Among all configurations, the 4-model decorrelated ensemble (Deco-LITETime-4) achieves the highest mean accuracy even surpassing the state-of-the-art LITETime-5 which is an ensemble of five LITE classifiers.
This performance clearly demonstrate efficiency of decorrelated learning in extracting complementary features which enables superior classification accuracy with fewer models.
The $p$-value between the Deco-LITETime-4 and LITETime-5 indicates that these two classifiers are not statistically different.
In contrast the $p$-value between LITETime-4 and LITETime-5 is lower than 5\% which highlights a statistically significant difference.
From the figure we can also notice that Deco-LITETime-4 also outperform its couterpart, LITETime-4 with a statistically significant lower $p$-value.
Furthermore, the results from the Figure~\ref{fig:mcm_sota} also indicate that the Deco-LITETime-4 performs almost on par with LITETime-5 despite using one fewer model. 
Additionally, a one-vs-one performance comparison between Deco-LITETime-4 and LITETime-5 is shown in the Figure~\ref{fig:dec_ens_4_vs_litetime}.

Similarly, from the Figure~\ref{fig:mcm_sota}, we can also  observe that the 3-model decorrelated ensemble (Deco-LITETime-3) achieves comparable performance to the more parameter-intensive LITETime-5. 
The difference between Deco-LITETime-3 and LITETime-5 is not statistically significant, as indicated by the high $p$-value. 
The performance of Deco-LITETime-3 establishes it as a viable alternative to the SOTA LITETime-5 whille offering comparable results with two fewer models.

From the same figure, it seems that advantages of decorrelated learning are less evident in smaller ensembles.
The 2-model decorrelated ensemble (Deco-LITETime-2) shows only a marginal improvement over its base counterpart (LITETime-2) and does not fuly bridge performance gap with LITETime-5.
These results shows that benefits of decorrelated learning become more apparent as ensemle size increases.
It can be explained based on the fact that, in smaller ensembles, features from 2-base models already exhibit some inherent diversity which diminishes immediate impact of decorrelation.
The feature overlap among base models increases as the ensemble size grows in which decorrelated learning plays a crucial role in maximizing feature diversity and enahncing generalization and overall performance.

\begin{figure*}
    \centering
    \includegraphics[width=\linewidth]{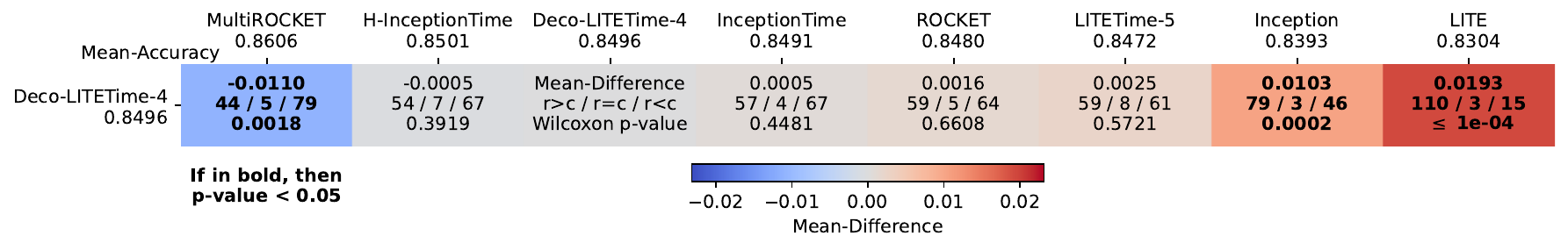}
    \caption{The Multi-Comparison Matrix applied to show the performance of Deco-LITETime-4 compared to state-of-the-art approaches.}
    \label{fig:comparison_sota}
\end{figure*} 

\subsection{Comparison with State-of-the-Art Methods}
As shown in Figure~\ref{fig:comparison_sota}, our proposed method achieves an average accuracy of 0.8496, ranking third among the evaluated models. 
It performs competitively against other SOTA deep learning approaches and surpasses several existing methods. 
The only approach that significantly outperforms it with statistical significance is MultiROCKET~\cite{tan2022multirocket} which remains the highest-ranked model. 
This highlights the effectiveness of our diversity-driven ensemble strategy in improving classification performance while maintaining efficiency. 
Notably, the Deco-LITETime-4 contains less than 10\% of the parameters of a single Inception-based classifier, demonstrating that our method achieves strong performance with significantly reduced computational cost. 
It is important to note that while the results for our baseline were obtained through our own experiments, the results for the other methods, including MultiROCKET, were sourced from the official repository~\cite{tan2022multirocket}. These findings further demonstrate that explicitly promoting feature diversity within ensembles enhances generalization and provides a strong alternative to conventional ensemble learning techniques.

\subsection{Quantitative Diversity Analysis}
\begin{figure}
    \centering
    \includegraphics[width=\linewidth]{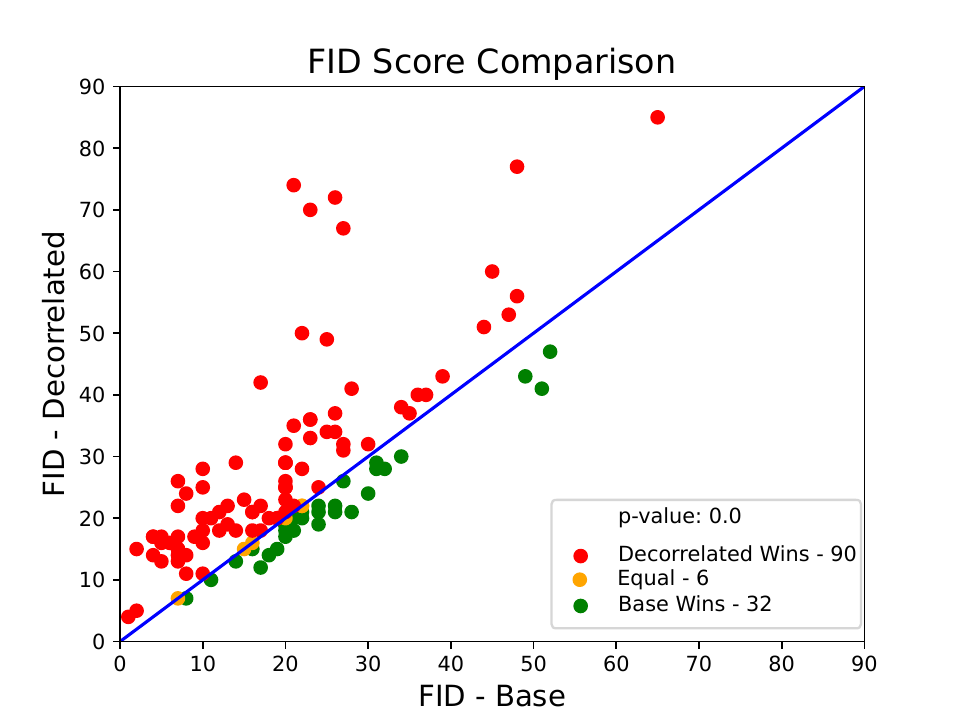}
    \caption{\centering FID score comparison between two base models and a base-decorrelated model pair.}
    \label{fig:fid_base_deco}
\end{figure}

To quantitatively assess the impact of our decorrelated learning framework on feature diversity, we compare two different ensemble configurations using the Fréchet Inception Distance (FID)~\cite{heusel2017gans,ismail2025establishing}. 
FID measures the similarity between two distributions where each distribution represents the statistical properties of features extracted by an individual model. 
By computing FID scores within two distinct 2-model ensembles, we can directly compare the feature diversity of base models and their decorrelated counterparts.
Figure~\ref{fig:fid_base_deco} presents a one-vs-one FID score comparison across 128 datasets from the UCR archive. 
In this figure, the x-axis represents the FID scores computed between a reference model and a base model while the y-axis represents the FID scores computed between the same reference model and corresponding decorrelated model.
The results indicate that the decorrelated model produces higher FID scores in 90 datasets, often with a larger margin while the base model results in higher FID scores in 32 datasets. 
In 6 datasets, there is no observed difference between the two configurations. 
The p-value computed between these two sets of FID scores is 0.0, confirming that the observed difference is statistically significant.
These findings demonstrate that decorrelated learning explicitly enhances feature diversity, encouraging models to learn more distinct representations. 
By reducing feature redundancy, this approach contributes to better generalization and improved ensemble performance in time series classification tasks.

\subsection{Qualitative Diversity Analysis}  
\begin{figure}[h!]
    \centering
    \includegraphics[width=0.9\linewidth]{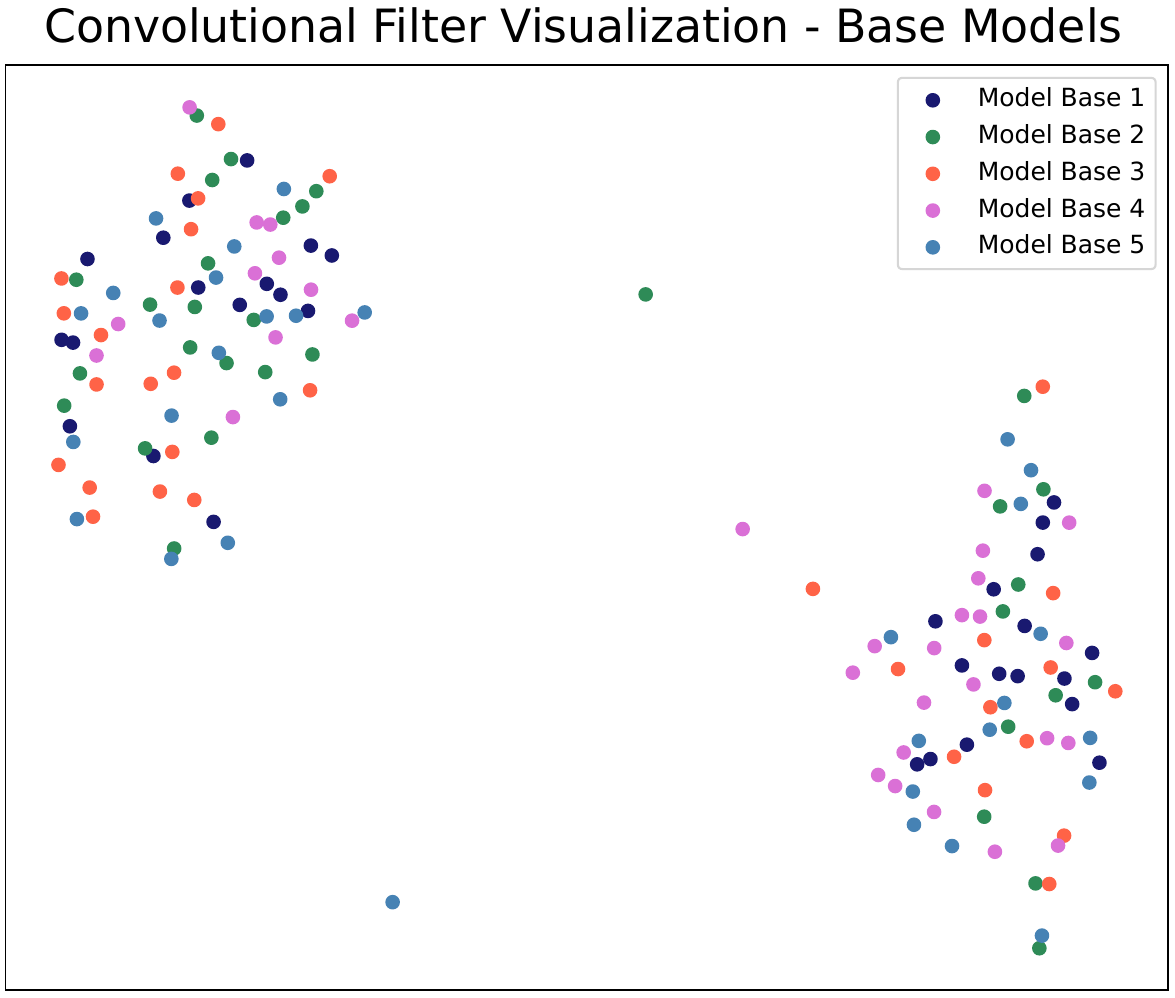}
    \caption{\centering t-SNE visualization of learned convolutional filters by highlighting redundancy in the filters of base models.}
    \label{fig:filters_base}
\end{figure}

To further emphasize our results, we analyze the filter space of base and decorrelated models. 
The primary objective of the decorrelated loss is to explicitly enhance feature diversity which indirectly drives convolutional filters to learn more distinct representations. 
To investigate this effect we visualize the learned convolutional filters from both base and decorrelated models to assess filter diversity.  
Following the default configuration outlined in the LITE model~\cite{ismail2023lite}, the number of filters in the final layer is 32, resulting in a convolutional filter dimension of 32 × 20 for each base and decorrelated model. 
To quantitatively analyze filter diversity we employ Dynamic Time Warping (DTW)~\cite{petitjean2014dynamic} to measure the similarity between all pairs of filters. 
Additionally, we employ t-distributed Stochastic Neighbor Embedding (t-SNE)~\cite{van2008visualizing} to project the high-dimensional filter representations into a two-dimensional space, enabling straightforward visualization within a Cartesian coordinate system.  
As illustrated in Figure~\ref{fig:filters_base}, the convolutional filters from all five base models exhibit a highly similar distribution, forming two distinct clusters where each cluster contains filters from all base models. 
This redundancy in the filter space clearly indicates that, despite different initializations, base models tend to learn highly similar filters.  

\begin{figure}[h!]
    \centering
    \includegraphics[width=0.9\linewidth]{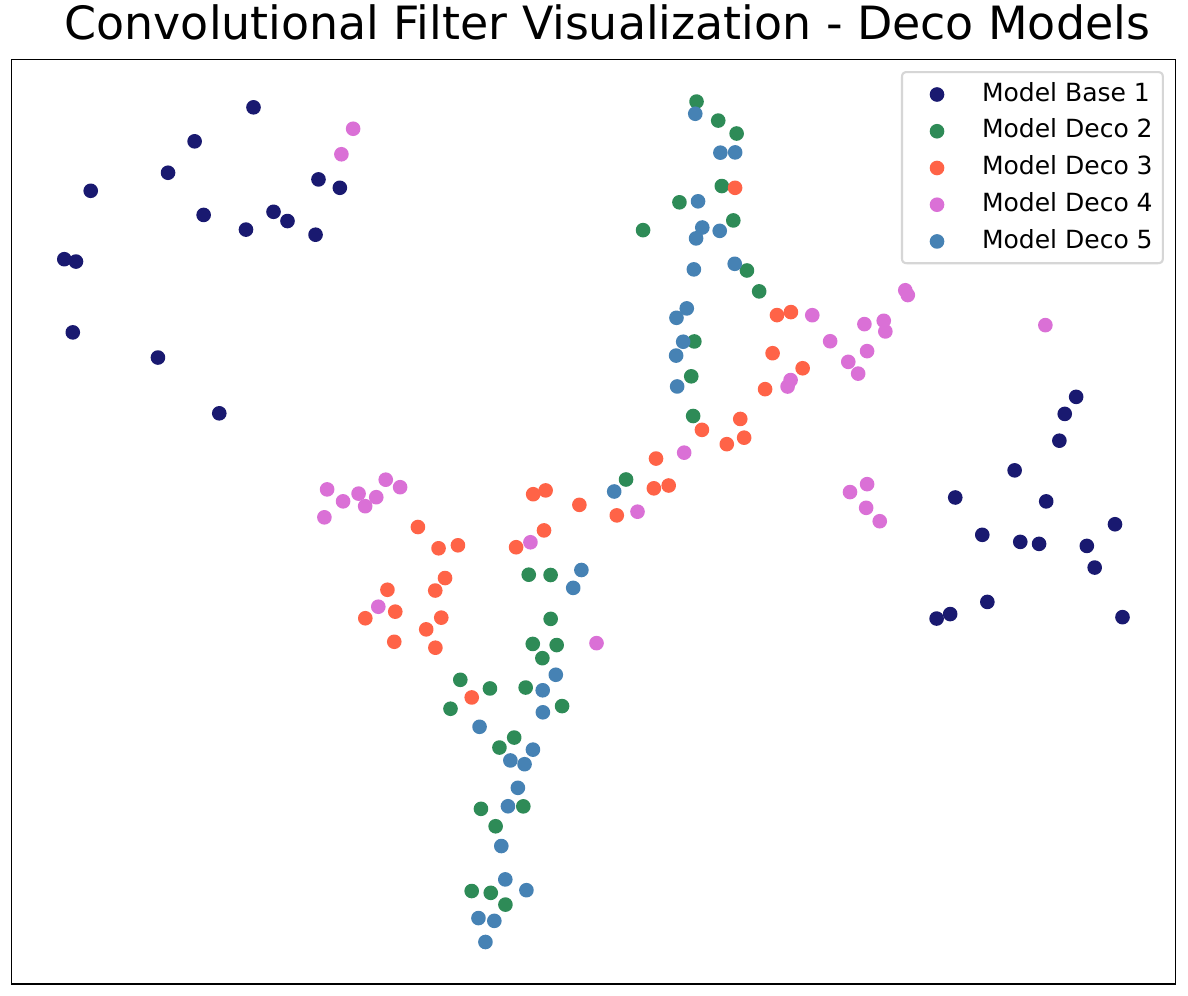}
    \caption{\centering t-SNE visualization of learned convolutional filters by highlighting diversity between the filters of base and decorrelated models.}
    \label{fig:filters_deco}
\end{figure}

In addition, we analyze the filter space of decorrelated models as shown in Figure~\ref{fig:filters_deco}. 
This figure includes convolutional filters from the first base model (used as a reference model) along with all four decorrelated models. 
The diversity between base and decorrelated filters is evident and the decorrelated models themselves also exhibit diversity among them.  
As previously mentioned, the decorrelation loss for the \(n\)-th model is computed as the sum of individual diversity losses with each previously trained model \((1, ..., n-1)\). 
From the results it seems that the diversity loss between a given decorrelated model and the base model is easier to optimize than that of previously trained decorrelated models.
We believe that fine-tuning the weighting of individual diversity losses within the total diversity loss can lead to better results as the optimal trade-off may vary across datasets.  
From Figure~\ref{fig:filters_deco}, we can observe that the filters of Deco\_5 are distributed similarly to those of Deco\_2, suggesting that the fifth decorrelated model fails to introduce additional diversity. 
This can be explained by the fact that most diverse and useful features have already been captured by previously trained decorrelated models, leaving Deco\_5 with limited room to learn new and distinctive patterns.  
Furthermore, it is also worth noting that the 2-model decorrelated ensemble (Deco-LITETime-2) achieves approximately 3.6\% higher test accuracy than LITETime-5 which is an ensemble of five base models. 
These results highlight the effectiveness of our decorrelated learning framework, demonstrating that smaller but more diverse ensembles can outperform larger ensembles composed of redundant models.

\section{Conclusion and Future Work}\label{sec:conclusion}
In this work, we proposed a diversity-driven ensemble learning framework for time series classification that explicitly encourages feature diversity through feature orthogonality loss. 
By enforcing decorrelation at the feature representation level, our method mitigates redundancy among ensemble members and improves generalization without requiring additional model complexity. 
Comprehensive experiments on 128 datasets from the UCR archive demonstrated that our approach achieves SOTA performance with fewer models, highlighting the efficiency of diversity-driven ensembling. 
Both quantitative and qualitative analyses confirmed that enforcing feature diversity results in more complementary feature representations, leading to improved classification accuracy.
Future work will explore refining the balance between classification loss and diversity loss, optimizing the trade-off dynamically across different datasets. 
Additionally, we aim to extend the framework to multivariate time series classification and investigate alternative diversity-promoting strategies beyond feature orthogonality. 
Further research will also focus on improving computational efficiency by potentially leveraging parallel training strategies to enhance scalability.

\section*{Acknowledgment}
This work was supported by the ANR DELEGATION project (grant ANR-21-CE23-0014) of the French Agence Nationale de la Recherche. 
The authors would like to acknowledge the High Performance Computing Center of the University of Strasbourg for supporting this work by providing scientific support and access to computing resources. 
Part of the computing resources were funded by the Equipex Equip@Meso project (Programme Investissements d’Avenir) and the CPER Alsacalcul/Big Data. 
The authors would also like to thank the creators and providers of the UCR Archive.

\bibliographystyle{IEEEtran} % Use the IEEE style
\bibliography{reference}    % Point to your .bib file

@article{dau2018optimizing,
  title={Optimizing dynamic time warping’s window width for time series data mining applications},
  author={Dau, Hoang Anh and Silva, Diego Furtado and Petitjean, Fran{\c{c}}ois and Forestier, Germain and Bagnall, Anthony and Mueen, Abdullah and Keogh, Eamonn},
  journal={Data mining and knowledge discovery},
  volume={32},
  pages={1074--1120},
  year={2018},
  publisher={Springer}
}

@article{mohammadi2024deep,
  title={Deep learning for time series classification and extrinsic regression: A current survey},
  author={Mohammadi Foumani, Navid and Miller, Lynn and Tan, Chang Wei and Webb, Geoffrey I and Forestier, Germain and Salehi, Mahsa},
  journal={ACM Computing Surveys},
  volume={56},
  number={9},
  pages={1--45},
  year={2024},
  publisher={ACM New York, NY}
}

@article{rajkomar2018scalable,
  title={Scalable and accurate deep learning with electronic health records},
  author={Rajkomar, Alvin and Oren, Eyal and Chen, Kai and Dai, Andrew M and Hajaj, Nissan and Hardt, Michaela and Liu, Peter J and Liu, Xiaobing and Marcus, Jake and Sun, Mimi and others},
  journal={NPJ digital medicine},
  volume={1},
  number={1},
  pages={18},
  year={2018},
  publisher={Nature Publishing Group UK London}
}

@article{nweke2018deep,
  title={Deep learning algorithms for human activity recognition using mobile and wearable sensor networks: State of the art and research challenges},
  author={Nweke, Henry Friday and Teh, Ying Wah and Al-Garadi, Mohammed Ali and Alo, Uzoma Rita},
  journal={Expert Systems with Applications},
  volume={105},
  pages={233--261},
  year={2018},
  publisher={Elsevier}
}

@article{pelletier2019temporal,
  title={Temporal convolutional neural network for the classification of satellite image time series},
  author={Pelletier, Charlotte and Webb, Geoffrey I and Petitjean, Fran{\c{c}}ois},
  journal={Remote Sensing},
  volume={11},
  number={5},
  pages={523},
  year={2019},
  publisher={MDPI}
}

@inproceedings{yi2018integrated,
  title={An integrated model for crime prediction using temporal and spatial factors},
  author={Yi, Fei and Yu, Zhiwen and Zhuang, Fuzhen and Zhang, Xiao and Xiong, Hui},
  booktitle={2018 IEEE International Conference on Data Mining (ICDM)},
  pages={1386--1391},
  year={2018},
  organization={IEEE}
}

@article{lucas2019proximity,
  title={Proximity forest: an effective and scalable distance-based classifier for time series},
  author={Lucas, Benjamin and Shifaz, Ahmed and Pelletier, Charlotte and O’Neill, Lachlan and Zaidi, Nayyar and Goethals, Bart and Petitjean, Fran{\c{c}}ois and Webb, Geoffrey I},
  journal={Data Mining and Knowledge Discovery},
  volume={33},
  number={3},
  pages={607--635},
  year={2019},
  publisher={Springer}
}

@inproceedings{bagheri2016support,
  title={Support vector machines with time series distance kernels for action classification},
  author={Bagheri, Mohammad Ali and Gao, Qigang and Escalera, Sergio},
  booktitle={2016 IEEE Winter Conference on Applications of Computer Vision (WACV)},
  pages={1--7},
  year={2016},
  organization={IEEE}
}

@article{ismail2019deep,
  title={Deep learning for time series classification: a review},
  author={Ismail Fawaz, Hassan and Forestier, Germain and Weber, Jonathan and Idoumghar, Lhassane and Muller, Pierre-Alain},
  journal={Data mining and knowledge discovery},
  volume={33},
  number={4},
  pages={917--963},
  year={2019},
  publisher={Springer}
}

@inproceedings{wang2017time,
  title={Time series classification from scratch with deep neural networks: A strong baseline},
  author={Wang, Zhiguang and Yan, Weizhong and Oates, Tim},
  booktitle={2017 International joint conference on neural networks (IJCNN)},
  pages={1578--1585},
  year={2017},
  organization={IEEE}
}

@article{ismail2020inceptiontime,
  title={Inceptiontime: Finding alexnet for time series classification},
  author={Ismail Fawaz, Hassan and Lucas, Benjamin and Forestier, Germain and Pelletier, Charlotte and Schmidt, Daniel F and Weber, Jonathan and Webb, Geoffrey I and Idoumghar, Lhassane and Muller, Pierre-Alain and Petitjean, Fran{\c{c}}ois},
  journal={Data Mining and Knowledge Discovery},
  volume={34},
  number={6},
  pages={1936--1962},
  year={2020},
  publisher={Springer}
}

@inproceedings{szegedy2017inception,
  title={Inception-v4, inception-resnet and the impact of residual connections on learning},
  author={Szegedy, Christian and Ioffe, Sergey and Vanhoucke, Vincent and Alemi, Alexander},
  booktitle={Proceedings of the AAAI conference on artificial intelligence},
  volume={31},
  number={1},
  year={2017}
}

@inproceedings{ismail2022deep,
  title={Deep learning for time series classification using new hand-crafted convolution filters},
  author={Ismail-Fawaz, Ali and Devanne, Maxime and Weber, Jonathan and Forestier, Germain},
  booktitle={2022 IEEE International Conference on Big Data (Big Data)},
  pages={972--981},
  year={2022},
  organization={IEEE}
}

@inproceedings{dietterich2000ensemble,
  title={Ensemble methods in machine learning},
  author={Dietterich, Thomas G},
  booktitle={International workshop on multiple classifier systems},
  pages={1--15},
  year={2000},
  organization={Springer}
}

@article{breiman1996bagging,
  title={Bagging predictors},
  author={Breiman, Leo},
  journal={Machine learning},
  volume={24},
  pages={123--140},
  year={1996},
  publisher={Springer}
}

@inproceedings{freund1996experiments,
  title={Experiments with a new boosting algorithm},
  author={Freund, Yoav and Schapire, Robert E and others},
  booktitle={icml},
  volume={96},
  pages={148--156},
  year={1996},
  organization={Citeseer}
}

@article{wolpert1992stacked,
  title={Stacked generalization},
  author={Wolpert, David H},
  journal={Neural networks},
  volume={5},
  number={2},
  pages={241--259},
  year={1992},
  publisher={Elsevier}
}

@article{kuncheva2003measures,
  title={Measures of diversity in classifier ensembles and their relationship with the ensemble accuracy},
  author={Kuncheva, Ludmila I and Whitaker, Christopher J},
  journal={Machine learning},
  volume={51},
  pages={181--207},
  year={2003},
  publisher={Springer}
}

@inproceedings{lee2023importance,
  title={On the importance of feature decorrelation for unsupervised representation learning in reinforcement learning},
  author={Lee, Hojoon and Lee, Koanho and Hwang, Dongyoon and Lee, Hyunho and Lee, Byungkun and Choo, Jaegul},
  booktitle={International Conference on Machine Learning},
  pages={18988--19009},
  year={2023},
  organization={PMLR}
}

@article{lines2015time,
  title={Time series classification with ensembles of elastic distance measures},
  author={Lines, Jason and Bagnall, Anthony},
  journal={Data Mining and Knowledge Discovery},
  volume={29},
  pages={565--592},
  year={2015},
  publisher={Springer}
}

@article{yang2020orthogonality,
  title={Orthogonality loss: Learning discriminative representations for face recognition},
  author={Yang, Shanming and Deng, Weihong and Wang, Mei and Du, Junping and Hu, Jiani},
  journal={IEEE Transactions on Circuits and Systems for Video Technology},
  volume={31},
  number={6},
  pages={2301--2314},
  year={2020},
  publisher={IEEE}
}

@article{tumer1996error,
  title={Error correlation and error reduction in ensemble classifiers},
  author={Tumer, Kagan and Ghosh, Joydeep},
  journal={Connection science},
  volume={8},
  number={3-4},
  pages={385--404},
  year={1996},
  publisher={Taylor \& Francis}
}

@article{ayinde2019regularizing,
  title={Regularizing deep neural networks by enhancing diversity in feature extraction},
  author={Ayinde, Babajide O and Inanc, Tamer and Zurada, Jacek M},
  journal={IEEE transactions on neural networks and learning systems},
  volume={30},
  number={9},
  pages={2650--2661},
  year={2019},
  publisher={IEEE}
}

@inproceedings{ismail2023lite,
  title={Lite: Light inception with boosting techniques for time series classification},
  author={Ismail-Fawaz, Ali and Devanne, Maxime and Berretti, Stefano and Weber, Jonathan and Forestier, Germain},
  booktitle={2023 IEEE 10th International Conference on Data Science and Advanced Analytics (DSAA)},
  pages={1--10},
  year={2023},
  organization={IEEE}
}

@article{yosinski2014transferable,
  title={How transferable are features in deep neural networks?},
  author={Yosinski, Jason and Clune, Jeff and Bengio, Yoshua and Lipson, Hod},
  journal={Advances in neural information processing systems},
  volume={27},
  year={2014}
}

@article{zhang2019nonlinear,
  title={Nonlinear regression via deep negative correlation learning},
  author={Zhang, Le and Shi, Zenglin and Cheng, Ming-Ming and Liu, Yun and Bian, Jia-Wang and Zhou, Joey Tianyi and Zheng, Guoyan and Zeng, Zeng},
  journal={IEEE transactions on pattern analysis and machine intelligence},
  volume={43},
  number={3},
  pages={982--998},
  year={2019},
  publisher={IEEE}
}

@inproceedings{wang2020orthogonal,
  title={Orthogonal convolutional neural networks},
  author={Wang, Jiayun and Chen, Yubei and Chakraborty, Rudrasis and Yu, Stella X},
  booktitle={Proceedings of the IEEE/CVF conference on computer vision and pattern recognition},
  pages={11505--11515},
  year={2020}
}

@article{dau2019ucr,
  title={The UCR time series archive},
  author={Dau, Hoang Anh and Bagnall, Anthony and Kamgar, Kaveh and Yeh, Chin-Chia Michael and Zhu, Yan and Gharghabi, Shaghayegh and Ratanamahatana, Chotirat Ann and Keogh, Eamonn},
  journal={IEEE/CAA Journal of Automatica Sinica},
  volume={6},
  number={6},
  pages={1293--1305},
  year={2019},
  publisher={IEEE}
}

@article{ismail2023approach,
  title={An approach to multiple comparison benchmark evaluations that is stable under manipulation of the comparate set},
  author={Ismail-Fawaz, Ali and Dempster, Angus and Tan, Chang Wei and Herrmann, Matthieu and Miller, Lynn and Schmidt, Daniel F and Berretti, Stefano and Weber, Jonathan and Devanne, Maxime and Forestier, Germain and others},
  journal={arXiv preprint arXiv:2305.11921},
  year={2023}
}

@article{benavoli2016should,
  title={Should we really use post-hoc tests based on mean-ranks?},
  author={Benavoli, Alessio and Corani, Giorgio and Mangili, Francesca},
  journal={The Journal of Machine Learning Research},
  volume={17},
  number={1},
  pages={152--161},
  year={2016},
  publisher={JMLR. org}
}

@incollection{wilcoxon1992individual,
  title={Individual comparisons by ranking methods},
  author={Wilcoxon, Frank},
  booktitle={Breakthroughs in statistics: Methodology and distribution},
  pages={196--202},
  year={1992},
  publisher={Springer}
}

@article{middlehurst2024bake,
  title={Bake off redux: a review and experimental evaluation of recent time series classification algorithms},
  author={Middlehurst, Matthew and Sch{\"a}fer, Patrick and Bagnall, Anthony},
  journal={Data Mining and Knowledge Discovery},
  pages={1--74},
  year={2024},
  publisher={Springer}
}

@article{middlehurst2021hive,
  title={HIVE-COTE 2.0: a new meta ensemble for time series classification},
  author={Middlehurst, Matthew and Large, James and Flynn, Michael and Lines, Jason and Bostrom, Aaron and Bagnall, Anthony},
  journal={Machine Learning},
  volume={110},
  number={11},
  pages={3211--3243},
  year={2021},
  publisher={Springer}
}

@inproceedings{badi2024cocalite,
  title={COCALITE: A Hybrid Model COmbining CAtch22 and LITE for Time Series Classification},
  author={Badi, Oumaima and Devanne, Maxime and Ismail-Fawaz, Ali and Abdullayev, Javidan and Lemaire, Vincent and Berretti, Stefano and Weber, Jonathan and Forestier, Germain},
  booktitle={2024 IEEE International Conference on Big Data (BigData)},
  pages={1229--1236},
  year={2024},
  organization={IEEE}
}

@article{van2008visualizing,
  title={Visualizing data using t-SNE.},
  author={Van der Maaten, Laurens and Hinton, Geoffrey},
  journal={Journal of machine learning research},
  volume={9},
  number={11},
  year={2008}
}

@article{heusel2017gans,
  title={Gans trained by a two time-scale update rule converge to a local nash equilibrium},
  author={Heusel, Martin and Ramsauer, Hubert and Unterthiner, Thomas and Nessler, Bernhard and Hochreiter, Sepp},
  journal={Advances in neural information processing systems},
  volume={30},
  year={2017}
}

@inproceedings{petitjean2014dynamic,
  title={Dynamic time warping averaging of time series allows faster and more accurate classification},
  author={Petitjean, Fran{\c{c}}ois and Forestier, Germain and Webb, Geoffrey I and Nicholson, Ann E and Chen, Yanping and Keogh, Eamonn},
  booktitle={2014 IEEE international conference on data mining},
  pages={470--479},
  year={2014},
  organization={IEEE}
}

@inproceedings{ay2022kdfcn,
  author = {Ay, E. and Devanne, M. and Weber, J. and Forestier, G.},
  title = {A study of Knowledge Distillation in Fully Convolutional Network for Time Series Classification},
  booktitle = {International Joint Conference on Neural Networks (IJCNN)},
  city = {Padua},
  country = {Italy},
  pages = {1--8},
  url = {https://doi.org/10.1109/IJCNN55064.2022.9892915},
  year = {2022},
  organization = {IEEE}
}

@article{ismail2025establishing,
  title={Establishing a unified evaluation framework for human motion generation: A comparative analysis of metrics},
  author={Ismail-Fawaz, Ali and Devanne, Maxime and Berretti, Stefano and Weber, Jonathan and Forestier, Germain},
  journal={Computer Vision and Image Understanding},
  volume={254},
  pages={104337},
  year={2025},
  publisher={Elsevier}
}

@article{tan2022multirocket,
  title={MultiRocket: multiple pooling operators and transformations for fast and effective time series classification},
  author={Tan, Chang Wei and Dempster, Angus and Bergmeir, Christoph and Webb, Geoffrey I},
  journal={Data Mining and Knowledge Discovery},
  volume={36},
  number={5},
  pages={1623--1646},
  year={2022},
  publisher={Springer}
}

\end{document}